\documentclass[sn-basic,iicol]{sn-jnl}

\usepackage{graphicx}%
\usepackage{multirow}%
\usepackage{amsmath,amssymb,amsfonts}%
\usepackage{amsthm}%
\usepackage{mathrsfs}%
\usepackage[title]{appendix}%
\usepackage{xcolor}%
\usepackage{textcomp}%
\usepackage{manyfoot}%
\usepackage{booktabs}%
\usepackage{algorithm}%
\usepackage{algorithmicx}%
\usepackage{algpseudocode}%
\usepackage[algo2e,inoutnumbered,algoruled,vlined]{algorithm2e}

\usepackage{listings}%

\def\msx{\mathsf{X}}

\def\msz{\mathsf{Z}}

\newcommand{\mcb}[1]{\mathcal{B}(#1)}

\def\mcx{\mathcal{X}}

\def\mcf{\mathcal{F}}
\def\mcg{\mathcal{G}}
\def\mch{\mathcal{H}}

\def\mcq{\mathcal{Q}}
\def\mct{\mathcal{T}}
\def\mcn{\mathcal{N}}

\def\mcn{\mathcal{N}}

\def\rset{\mathbb{R}}

\def\rmd{\mathrm{d}}

\newcommand{\floor}[1]{\left\lfloor #1 \right\rfloor}

\newcommand{\PE}{\mathbb{E}}

\newcommand{\parenthese}[1]{\left(#1 \right)}

\newcommand{\parentheseDeux}[1]{\left[ #1 \right]}

\newcommand{\expeMarkov}[2]{\PE_{#1} \left[ #2 \right]}

\def\ie{\textit{i.e.}}

\newcommand{\opnorm}[1]{{\left\vert\kern-0.25ex\left\vert\kern-0.25ex\left\vert #1 
    \right\vert\kern-0.25ex\right\vert\kern-0.25ex\right\vert}}

\def\I{\operatorname{I}}

\newcommand\coupling[2]{\Gamma(\mu,\nu)}

\renewcommand{\geq}{\geqslant}
\renewcommand{\leq}{\leqslant}

\def\KL{\mathrm{KL}}

\begin{document}

\title[Learning variational autoencoders via MCMC speed measures]{Learning variational autoencoders via MCMC speed measures}


\author[1]{\fnm{Marcel} \sur{Hirt}}
\equalcont{These authors contributed equally to this work.}

\author[2]{\fnm{Vasileios} \sur{Kreouzis}}
 \equalcont{These authors contributed equally to this work.}

\author*[2,3]{\fnm{Petros} \sur{Dellaportas}}\email{p.dellaportas@ucl.ac.uk}

\affil[1]{\orgdiv{School of Social Sciences \& School of Physical and Mathematical Sciences}, \orgname{Nanyang Technological University}, 
\country{Singapore}}

\affil*[2]{\orgdiv{Department of Statistical Science}, \orgname{University College London}, 
\country{UK}}

\affil[3]{\orgdiv{Department of Statistics}, \orgname{Athens University of Economics and Business},  \country{Greece}}


\abstract{Variational autoencoders (VAEs) are popular likelihood-based generative models which can be efficiently trained by maximizing an Evidence Lower Bound (ELBO). There has been much progress in improving the expressiveness of the variational distribution to obtain tighter variational bounds and increased generative performance. Whilst previous work has leveraged Markov chain Monte Carlo (MCMC) methods for the construction of variational densities, gradient-based methods for adapting the proposal distributions for deep latent variable models have received less attention. This work suggests an entropy-based adaptation for a short-run Metropolis-adjusted Langevin (MALA) or Hamiltonian Monte Carlo (HMC) chain while optimising a tighter variational bound to the log-evidence. Experiments show that this approach yields higher held-out log-likelihoods as well as improved generative metrics. Our implicit variational density can adapt to complicated posterior geometries of latent hierarchical representations arising in hierarchical VAEs.}




\maketitle

\section{Introduction}\label{sec1}

VAEs \citep{kingma2014auto, rezende2014stochastic} are powerful latent variable models that routinely use neural networks to parameterise conditional distributions of observations given a latent representation. This renders the Maximum-Likelihood Estimation (MLE) of such models intractable, so one commonly resorts to extensions of Expectation-Maximization (EM) approaches that maximize a lower bound on the data log-likelihood. These objectives introduce a variational or encoding distribution of the latent variables that approximates the true posterior distribution of the latent variable given the observation. However, VAEs have shortcomings; for example, they can struggle to generate high-quality images. These shortcomings have been attributed to failures to  match corresponding distributions in the latent space. First, the VAE prior can be significantly different from the aggregated approximate posterior \citep{hoffman2016elbo, rosca2018distribution}. To alleviate this prior hole phenomenon, previous work has considered more flexible priors, such as mixtures \citep{tomczak2017vae}, normalising flows \citep{kingma2016improved}, hierarchical priors \citep{klushyn2019learning} or energy-based models \citep{du2019implicit, aneja2021contrastive}. Second, the encoding distribution can be significantly different from the true posterior distribution. It has been an ongoing challenge to reduce this approximation error by constructing new flexible variational families \citep{barber1998ensemble, rezende2015variational}. 

This work utilises adaptive MCMC kernels to construct an implicit variational distribution, that, by the reversibility of the associated Markov kernel, decreases the Kullback-Leibler (KL) between an initial encoding distribution and the true posterior. In summary, this paper (i) develops gradient-based adaptive MCMC methods that give rise to flexible implicit variational densities for training VAEs; (ii) shows that non-diagonal preconditioning schemes are beneficial for learning hierarchical structures within VAEs; and (iii) illustrates the improved generative performance for different data sets and MCMC schemes. Our code is available at \url{https://github.com/kreouzisv/smvaes}.

\section{Background}

We are interested in learning deep generative latent variable models using VAEs.
Let $\msx \subset \rset^{d_x}$, $\msz \subset \rset^{d_z}$ and assume some \emph{prior} density $p_{\theta}(z)$ for $z \in \msz$, with all densities assumed with respect to the Lebesgue measure. The prior density can be fixed or made dependent on some parameters $\theta \in \Theta$. Consider a conditional density $p_{\theta}(x|z)$, also called \emph{decoder}, with $z \in \msz$, $x \in \msx$ and parameters also denoted $\theta$. We can interpret this decoder as a generative network that tries to explain a data point $x$ using a latent variable $z$. This latent structure yields the following distribution of the data
\[ p_{\theta}(x)= \int_{\msx} p_{\theta}(x|z)p_{\theta}(z)\rmd z.\]
Assume a ground truth measure $\mu$ on $\msx$, which can be seen as the empirical distribution of some observed data set. We want to minimize the negative log-likelihood with respect to $\mu$, \ie~ 
$ \min_{\theta \in \Theta} \int_{\mcx} -\log p_{\theta}(x) \mu (\rmd x)$.
Variational inference approaches approximately minimize the negative log-likelihood by maximizing the \emph{evidence lower bound} (ELBO)
\begin{align*}\mathcal{L}(\theta,\phi, x)=\expeMarkov{q_{\phi}(z|x)}{\log p_{\theta}(x|z)}- \KL(q_{\phi}(z|x)|p_{\theta}(z)) 
\end{align*}
averaged over $x \sim \mu$ and where $q_{\phi}(z|x)$ is a conditional distribution with parameter $\phi \in \Phi$, commonly termed \emph{encoder} and $\KL(q(z)|p(z))=\int_{\msz} q(z)\parenthese{\log q(z)-\log p(z)} \rmd z$ the Kullback-Leibler divergence between two densities $q$ and $p$. Consider also the \emph{posterior} density $p_{\theta}(z|x)\propto p_{\theta}(z) p_{\theta}(x|z)$. Then
\[\mathcal{L}(\theta,\phi,z)= -\log p_{\theta}(x) + \KL(q_{\phi}(z|x)|p_{\theta}(z|x).\]

\section{Related work} 
Many approaches have been proposed for combining MCMC with variational inference. \citet{salimans2015markov} and \citet{wolf2016variational} construct as a joint objective a variational bound on an extended state space that includes multiple samples of the Markov chain, extended in \citet{caterini2018hamiltonian} using tempering and illustrating connections with SMC samplers. Most related are the work \citet{hoffman2017learning, levy2018generalizing, hoffman2019neutra} that largely follow the same approach to estimate the generative parameters of the decoder, as well as the initial variational distribution, but differ in the way of adapting the Markov chain. \citet{titsias2017learning} suggested a model reparameterization using a transport mapping, while \citet{ruiz2019contrastive} suggested using a variational contrastive divergence instead of a KL divergence used herein. \citet{thin2020metflow} presented a variational objective on an extended space of the accept/reject variables that allows for entropy estimates of the distribution of the final state of the Markov chain. \citet{nijkamp2020learning} have used short-run MCMC approximations based on unadjusted Langevin samplers to train the generative model with extensions to learn energy-based priors in \citet{pang2020learning}. \citet{ruiz2021unbiased} used couplings for Markov chains to construct unbiased estimates of the marginal log-likelihood. 
Non-adaptive MCMC transitions have also been utilized in \citet{wu2020stochastic} to build stochastic normalising flows that approximates the posterior distribution in VAEs, but are trained by minimizing a KL-divergence between the forward and backward path probabilities, see also \citet{hagemann2022stochastic}.
More recently, \citet{taniguchi2022langevin} considered an amortised energy function over the encoder parameters and used a MALA algorithm to sample from its invariant distribution. \citet{peis2022missing} learn an initial encoding distribution based on a sliced Kernel Stein Discrepancy and then apply a non-adapted HMC algorithm.

\section{Training VAEs with MCMC speed measures}
\paragraph{Markov kernels.}	
We fix $x \in \msx$ and consider now a Markov kernel $M^k_{\theta,\phi_k}(\cdot |x)$ that can depend on $x$ with parameters $\phi_k \in \Phi_k$ and that shares the parameter $\theta\in \Theta$ from the generative model. Assume that the kernel is reversible with respect to $p_{\theta}(z|x)$. 
We can construct a distribution $q^K_{\theta,\phi}(z|x)$ by first sampling from an initial tractable distribution $q^0_{\phi_0}(z|x)$ and then applying the $K$  Markov kernels $M^k_{\theta,\phi_k}(\cdot|x)$ for $k\in \{1, \ldots , K\}$. Put another way, consider the following variational family
\begin{align*}\mcq_x =\{&q^K_{\theta,\phi}(\cdot |x)=q^0_{\phi_0}(\cdot |x) M^1_{\theta,\phi_1}(\cdot|x) \cdots  M^K_{\theta,\phi_K}(\cdot|x) ~,\\ & \phi_k \in \Phi_k, \phi=(\phi_0, \ldots, \phi_K), \theta \in \Theta\},\end{align*}
where $(qM)(z'|x)=\int_{\msz} q(z|x)M(z, z'|x) \rmd z$ for a conditional density $q(\cdot|x)$ and Markov kernel $M(\cdot|x)$ that depends on $x$. Although $q_{\theta,\phi}^K$ can be evaluated explicitly for the choice of Markov kernels considered here \citep{thin2020metflow}, we do not require this. Instead, we rely on the fact \citep{ambrosio2005gradient} that due to the reversibility of the Markov kernels with respect to $p_{\theta}(z|x)$, it holds that
\begin{equation}
    \KL\parenthese{q^K_{\theta,\phi}(z |x)| p_{\theta}(z|x)}\leq \KL\parenthese{q^0_{\phi_0}(z|x)| p_{\theta}(z|x)} . \label{eq:KL_improvement}
\end{equation} 
The non-asymptotic convergence of the Markov chain in KL-divergence depends on the posterior distribution as well as on the specific MCMC algorithm used, see for example \citet{cheng2018convergence} for the MALA case under convexity assumptions. 

We consider first a standard ELBO
\begin{align}
	&{\mathcal{L}_0}(\theta,\phi_0, x) \label{eq:standard_elbo} \\=& \expeMarkov{q^0_{\phi_0}(z|x)}{\log p_{\theta}(x|z)} - \KL(q^0_{\phi_0}(z|x)|p(z))
    \nonumber
\end{align}

Relation \eqref{eq:KL_improvement} motivates to learn $\phi_0$ by maximizing ${\mathcal{L}_0}(\theta,\phi_0)$. Indeed, due to
\begin{align*}
\KL\parenthese{q^K_{\theta,\phi}(z |x)| p_{\theta}(z|x)}&\leq \KL\parenthese{q^0_{\phi_0,}(z |x)| p_{\theta}(z|x)} \\
&= \log p_{\theta}(x) -{\mathcal{L}_0}(\theta,\phi_0),\end{align*}
maximizing ${\mathcal{L}_0}(\theta,\phi_0)$ decreases the KL divergence between the variational density $q^K_{\theta,\phi}(\cdot |x)$ and the posterior density for fixed $\theta$ and $\phi_1$.

We also need to specify the Markov kernels. We use \emph{reparameterisable} Metropolis-Hastings kernels with the potential function $U_{\theta}(z|x)=-\log p_{\theta}(x|z)-\log p_{\theta}(z)$ and target $\pi_{\theta}(z)=p_{\theta}(z|x) \propto \exp(-U_{\theta}(z|x)$. More precisely, for $A \in \mcb{\msz}$, 
\begin{align*} M^k_{\theta,\phi_1}(z,A|x)= & \int_{\msz} \nu(\rmd v)  \Big[  \parenthese{1- \alpha(z,z')} \delta _{z} (A) \\ & +\alpha(z, z') \delta _{z'} (A)  \Big]_{z'=\mct_{\theta,\phi_1}(v|z,x)} \end{align*}
where $\alpha(z,z')$ is an \emph{acceptance rate} for moving from state $z$ to $z'$, $\mct_{\theta,\phi_1}(\cdot|z,x)$ is a \emph{proposal mapping} and $\nu$ is a parameter-free density over $\msz$. Although the different Markov kernels could have different parameters $\phi_k$ for $k \in \{1, \ldots, K\}$, we assume for simplicity that they all share the parameters $\phi_1$. 

\paragraph{Speed measure adaptation. }
For a random walk Markov chain with isotropic proposal density $r(z,z')=\mcn(z, \sigma^2 \I)$ at position $z$, the speed measure is defined as $\sigma^2 \times \alpha(z)$, where $a(z)=\int \alpha(z,z') r(z,z') \rmd z'$ is the average acceptance rate.
To encourage fast mixing for the Markov chain across all dimensions jointly, \citet{titsias2019gradient} suggested a generalisation of this speed measure that amounts to choosing the parameters $h$ and $C$ from the proposal so that the proposal has both high acceptance rates, but also a high entropy. We, therefore, aim to choose $\phi_1$ that approximately maximizes
\begin{align*}
	 \mcf(\phi_1, & z,x)  =   \\
  & \parentheseDeux{ \int_{\msz} \log \alpha (z,\mct_{\theta,\phi_1}(v|z,x)) \nu(\rmd v) + \beta \mch_{\theta,\phi_1}},
\end{align*}  
averaged over $(x,z)\sim \mu(x) q^0_{\phi_0}(z|x)$ where $\beta>0$ is some hyper-parameter that can be updated online to achieve a desirable average acceptance rate $\alpha^\star$.

\paragraph{MALA.}
Consider first a Metropolis Adjusted Langevin Algorithm (MALA). We assume that $\phi_1$ parameterises a non-singular matrix $C$, possibly dependent on $x$, which can be, for instance, a diagonal matrix or a Cholesky factor. In this case, we can write the proposed state $z'$ as
\begin{equation}z'=\mct_{\theta,\phi_1}(v|z,x) = z-\frac{h^2}{2}  CC^\top \nabla U_{\theta}(z|x)+h Cv 
\end{equation}
for some step size $h>0$ that is part of the parameter $\phi_1$ and where $v\sim \nu=\mcn(0,\I)$. The log-acceptance rate is
$\log a(z, z')=\min \{0, -\Delta(v,z,z')\}$ based on the energy error
\begin{align*} \Delta(v, & z,z') = U_{\theta}(z'|x)-U_{\theta}(z|x) -\frac{1}{2}{\Vert v \Vert}^2\\
&+ \frac{1}{2} {\Big \Vert   v-  \frac{h}{2} C \left\{\nabla U_{\theta}(z|x)+ \nabla U_{\theta} (z'|x) \right\} \Big \Vert}^2, 
\end{align*}
evaluated at $z'=\mct_{\theta,\phi_1}(v|z,x)$.
The proposal density of the Markov kernel 
$$r(z,z'|x) = \mcn \left(z-\frac{h^2}{2} CC^\top \nabla U_{\theta}(z|x), h^2CC^\top \right)$$ 
can be viewed as the pushforward density of $\mcn(0,\I)$ with respect to the transformation $\mct_{\theta,\phi_1}(v|z,x)$. Its entropy is
\begin{align*}
\mch_{\theta,\phi_1}&=-\int_{\msz} r_{\theta,\phi_1}(z,z'|x) \log r_{\theta,\phi_1}(z,z'|x) \rmd z'\\
&= \text{const} +\log |\det (hC)|, \end{align*}
which is constant for $z \in \msz$ in the standard MALA case, although it can depend on $x$ for MALA with state dependent proposals. 

\paragraph{HMC.} Consider next a Hamiltonian Monte Carlo Algorithm (HMC) based on a leapfrog or velocity Verlet integrator with $L$ steps \citep{hairer2003geometric, bou2018geometric}. We assume that $\phi_1$ parameterises a Cholesky factor matrix $C$ of the inverse \emph{mass matrix} $M^{-1}=CC^\top$. The proposed state $z'=q_L$ is commonly computed recursively for $\ell \in \{1, \ldots, L\}$ via
\begin{align*}
p_{\ell + \frac{1}{2}} &= p_{\ell} - \frac{1}{2} \nabla U_{\theta}(q_{\ell}|x) \\
q_{\ell+1} &= q_{\ell} + h M^{-1} p_{\ell + \frac{1}{2}} \\ 
p_{\ell+1} &= p_{\ell+\frac{1}{2}} - \frac{1}{2} \nabla U_{\theta}(q_{\ell+1}|x),
\end{align*}
where $p_{\ell}$ is a sequence of \emph{momentum} variables initialised at $p_0=C^{-\top}v$ for $v \sim \mcn(0,\I)$. It is possible \citep{livingstone2019geometric, durmus2017convergence} to write the proposed state $z'=\mct_{\theta, \phi_1}(v|z,x)$ in the representation
$$ z'=z-\frac{L}{2} CC^\top \nabla U_{\theta}(z|x)+LCv - CC^\top \Xi_{L}(v)$$
where $\Xi_L(v)=\sum_{\ell=1}^{L-1}(L-\ell) \nabla U_{\theta}(q_{\ell})$ is a weighted average of the potential energy gradients along the leapfrog trajectory. However, $\Xi_L$ is generally non-linear and evaluating the log-determinant of the Jacobian of $\mct_L$ scales poorly for high dimensional latent variables. We, therefore, consider the approximation suggested in \citet{hirt2021entropy} based on a local Gaussian assumption
\begin{align*}
&\log r_{\theta,\phi_1}(z,\mct_L(v)) \\ \approx & \log \nu(v) - d \log L - \log |\det C| \\ & - \log \left|\det \left( \I - \frac{L^2-1}{6} C^\top \nabla U_{\theta}(q_{\floor{ L/2}} |x) C \right) \right|, \end{align*}
where Russian roulette estimators \citep{behrmann2019invertible, chen2019residual} can be used to compute the last term.

\paragraph{Learning the generative model.}
Maximizing the log-likelihood function directly using
$$ \nabla_{\theta} \log p_{\theta}(x)=\int_{\msz} p_{\theta}(z|x) \nabla_{\theta} \log p_{\theta}(x,z) \rmd z$$
is usually intractable as it requires samples from $p_{\theta}(z|x)$. Instead, we use samples from an MCMC chain, as used previously, see, for instance, \citet{han2017alternating, hoffman2017learning, nijkamp2020learning}. More precisely, at iteration $t$, let $\theta^{(t)}$ and $\phi^{(t)}$ be the current estimate of the generative and variational parameters. The resulting gradient update then arises as the gradient of the following perturbation of the log-likelihood
\begin{align*}
    &\mcg(\theta, \theta^{(t)}, \phi^{(t)},x) \\
    =& \log p_{\theta}(x) - \KL(q^K_{\theta^{(t)},\phi^{(t)}}(z|x)|p_{\theta}(z|x))\\
    =& \int_{\msz} q^K_{\theta^{(t)},\phi^{(t)}}(z|x) \left[ \log p_{\theta}(x) + \log p_{\theta}(z|x) \right] \rmd z +\text{const} 
\end{align*}
for $x \in \msx$, see also \citet{pang2020learning,han2020projections}. Here, $\text{const}$ is the entropy of $ q^K_{\theta^{(t)},\phi^{(t)}}(z|x)$, which is independent of $\theta$ so that the gradient with respect to $\theta$ becomes
\begin{align*}  \int_{\msz} q^K_{\theta^{(t)},\phi^{(t)}}(z|x) \left[\nabla_{\theta} \log p_{\theta}(z) + \nabla_{\theta}\log p_{\theta}(x|z)  \right] \rmd z .\end{align*}

\paragraph{Algorithm.} Pseudo-code for the suggested algorithm is given in Algorithm \ref{alg:implementation} at a given iteration $t$, for illustration based on a mini-batch of size one. We have found that pre-training the decoder and encoder parameters $\theta$, respectively $\phi_0$, by optimizing the standard ELBO \eqref{eq:standard_elbo} before applying Algorithm \ref{alg:implementation} can decrease the overall training time. While we only consider MALA or HMC proposals in our experiments, other proposals with a tractable entropy, such as from \citep{li2020neural}, could be used analogously. 

\begin{algorithm}[ht]
\KwIn{Number of Metropolis-Hastings steps $K$, learning rates $\rho^1,\rho^2,\rho^3,\rho^4$, current parameters $\theta^{(t)}$, $\phi_0^{(t)}$, $\phi_1^{(t)}$, $\beta^{(t)}$ and target acceptance rate $\alpha^\star$.}
$~$ Sample $x \sim \mu$.\\
Sample $z_0 \sim q_{\xi_0^{(0)}}(\cdot|x)$ via reparameterisation.\\
Set $\widehat{\nabla}_{\phi_0^{(t)}}{\mathcal{L}_0} \\=  \nabla_{\phi_0^{(t)}}\log p_{\theta^{(t)}}(x|z_0) - \nabla_{\phi_0^{(t)}}\log q_{\phi_0^{(t)}}(z_0|x) $ \\
Set $\widehat{\nabla}_{\phi^{(t)}}\mcf=0$ and $\widehat{\alpha}=0.$\\
Set $U_{\theta^{(t)}}(z|x)=-\log p_{\theta^{(t)}}(x|z)-p_{\theta^{(t)}}(z)$ with corresponding Markov kernel $M_{\theta^{(t)},\phi_1^{(t)}}$. \\
\For{k=1 \KwTo K}{
Sample $z_k \sim M_{\theta^{(t)},\phi_1^{(t)}}(z_{k-1},\cdot|x)$ via  $v_k \sim \mcn(0,\I)$ and $z_k=\mct_{\theta^{(t)},\phi_1^{(t)}}(v_k|z_{k-1},x)$.\\
Set $\widehat{\alpha} \mathrel{{\scriptstyle+}} = 1$ if $z_k$ is accepted.\\
Set $\widehat{\nabla}_{\phi_1^{(t)}}\mcf \mathrel{{\scriptstyle+}}= \nabla_{\phi_1^{(t)}} \big[ \log \alpha_{\theta^{(t)}, \phi_1^{(t)}}(z_{k-1}, z_k) -  \beta^{(t)} \log r_{\phi_1^{(t)}}(z_{k-1}, z_k|x) \big] $.
}
$~$ Set \\$\widehat{\nabla}_{\theta^{(t)}}\mcg= \nabla_{\theta} \parentheseDeux{\log p_{\theta}(x|z_K) + \log p_{\theta}(z_K)} |_{\theta=\theta^{(t)}}$.	\\	
Perform parameter updates: 	\\
$\phi_0^{(t+1)}=\phi_0^{(t)}+ \rho^1 \widehat{\nabla}_{\phi_0^{(t)}}{\mathcal{L}_0}$\\
$\phi_1^{(t+1)}=\phi_1^{(t)}+ \rho^2 \widehat{\nabla}_{\phi_1^{(t)}}{\mathcal{F}_0}$\\
$\theta^{(t+1)}=\theta^{(t)}+ \rho^3 \widehat{\nabla}_{\theta^{(t)}}{\mathcal{G}}$\\
$\beta^{(t+1)}=\beta^{(t)} (1 + \rho^4 (\frac{\widehat{\alpha}}{K} - \alpha^\star)).$\\
\caption{Single training step for updating the generative model, initial encoding distribution and MCMC kernel.}
\label{alg:implementation}
\end{algorithm}

\section{Extension to hierarchical VAEs}

We consider top-down hierarchical VAE (hVAE) architectures. Such models can leverage multiple layers $L$ of latent variables $(z_1, \ldots, z_L)$, $z_\ell \in \rset^{n_{\ell}}$ by generating them in the same order in both the prior 
\begin{equation}
(z_1, \ldots, z_L) \sim  p_{\theta}(z_1) p_{\theta}(z_2|z_1) \cdots p_{\theta}(z_L|z_1, \ldots, z_{L-1})
\label{eq:prior_hvae} 
\end{equation}
as well as in the approximate posterior,
\begin{equation} (z_1, \ldots, z_L)|x \sim q_{\phi_0,\theta}(z_1|x) \cdots q_{\phi_0,\theta}(z_L|x, z_1, \ldots, z_{L-1}) 
\label{eq:variational_approx_hvae}
\end{equation}
cf. \citet{sonderby2016ladder, kingma2016improved, nijkamp2020learning, maaloe2019biva, vahdat2020nvae, child2021very}. More concretely, we consider a sequence of variables $d_{\ell} \in \rset^{n'_{\ell}}$ that are deterministic given $z_{\ell}$ and defined recursively as \begin{equation}
d_{\ell}=h_{\ell,\theta}(z_{\ell-1}, d_{\ell-1}) \label{eq:deterministic_recursion}
\end{equation}
for some neural network function $h_{\ell,\theta}$, where the $d_{\ell}$-argument is a possible skip connection in a residual architecture for $\ell>1$ and some constant $d_1$. Suppose further that we instantiate \eqref{eq:prior_hvae} in the form
\begin{equation}
z_{\ell} = \mu_{\ell,\theta}(d_{\ell}) + \sigma_{\ell,\theta}(d_{\ell}) \odot \epsilon_{\ell}\label{eq:gaussian_hierarchical_prior} 
\end{equation}
for some functions $\mu_{\ell,\theta}$ and $\sigma_{\ell,\theta}$ and $\epsilon_{\ell}$ are iid Gaussian random variables. This construction leads to the auto-regressive structure in the prior \eqref{eq:prior_hvae}. To describe the variational approximation in \eqref{eq:variational_approx_hvae}, we consider a bottom-up network that defines deterministic variables $d_{\ell}'\in \rset^{n'_{\ell}}$ recursively by setting $d'_{L+1}=x$ and $d'_{\ell}=h'_{\ell, \phi_0}(d'_{\ell+1})$ for $1 \leq \ell\leq L$ for functions $h'_{\ell,\phi_0}$. We assume a residual parameterisation \citep{vahdat2020nvae,vahdat2021score} for $q_{\phi_0}(z_{\ell}|x, z_{<\ell})$ in the form
\begin{align} 
z_{\ell}=&\mu_{\ell,\theta}(d_{\ell}) + \sigma_{\ell,\theta}(d_{\ell}) \mu'_{\ell,\phi}(d_{\ell},d'_{\ell}) \nonumber\\&+ (\sigma_{\ell,\theta}(d_{\ell}) \sigma'_{\ell,\phi_0}(d_{\ell},d'_{\ell}) )  \odot \epsilon_{\ell}\label{eq:residual_param}
\end{align}
for some functions $\mu'_{\ell,\phi_0}$ and $\sigma'_{\ell, \phi_0}$. This implies that
\begin{align}
&\KL (q_{\phi_0}(z_\ell|x,z_{<\ell})|p_{\theta}(z_\ell|z_{<\ell})) \label{eq:hvae_kl}\\
=& \frac{1}{2} \Bigg[\sum_{i=1}^{n^\ell} \sigma_{\ell, \phi_0}'(d_\ell, d'_\ell)_i ^2 - n^\ell +  \mu'_{\ell,\phi}(d_{\ell},d'_{\ell})_i^2 \nonumber\\&+ \log  \sigma_{\ell, \phi_0}'(d_\ell, d'_\ell)_i^2 \Bigg]. \nonumber
\end{align}
The observations $x$ are assumed to depend explicitly only on $z_L$ and $d_L$ through  some function $g_{\theta}$ in the sense that $x|z_1, \ldots, z_L \sim p_{\theta}(x| g_{\theta}(z_L))$. The generative model of the latent variables $z_1, \ldots z_L$  in \eqref{eq:prior_hvae} is written in a centred parameterisation that makes them dependent a-priori. Our experiments will illustrate that these dependencies can make it sampling from the posterior difficult for MCMC schemes that are not adaptive.

HVAEs can be interpreted as diffusion discretisations \citep{falck2022multi}. Besides, diffusion models can be trained with a variational bound wherein the KL terms \eqref{eq:hvae_kl} at the intermediate layers are replaced by a denoising matching term \citep{ho2020denoising,kingma2021variational}. Note that the drift term in the denoising Markov chain in a score-based generative modelling formulation \citep{song2019generative,song2020score} of such models is based on an approximation of the (Stein) score function $ \nabla_x \log p_{\theta,t}(x)$, where $p_{\theta,t}$ is the density of the diffusion process at time $t$. Latent diffusion models \citep{rombach2022high} apply deterministic encoders to encode $x$ into a latent representation $z$, deterministic decoders to reconstruct $x$ from $z$, and consider a denoising diffusion framework for the latent represenations.
In contrast, the Markov chains in this work are based on the gradient of the transformed joint posterior density function $ CC^\top\nabla_z \log p_{\theta}(z_1, \ldots z_L|x)$. 



\section{Numerical Experiments}

\subsection{Evaluating Model Performance with Marginal log Likelihood}

We start by considering different VAE models and inference strategies on four standard image data sets (MNIST, Fashion-MNIST, Omniglot and SVHN) and evaluate their performance in terms of their test log-likelihood estimates.

\begin{table*}[t]
\caption{Importance sampling estimate of the log-likelihood on the test set based on $S=10000$ and $\tau = 1.5$. The values denote the mean of three independent runs, while the standard deviation is given within brackets. The MoG and VAMP VAEs use different priors.}
\label{log-likelihood-vae}
\vskip 0.15in
\begin{center}
\begin{small}
\begin{sc}
\begin{tabular}{lcccr}
\hline
Model & MNIST  & Fashion-MNIST  & Omniglot  & SVHN \\
\hline
VAE    & -81.16 (0.2)& -116.65 (0.1) & -117.46 (0.2) & 7.203 (0.005)\\
VAE-gradMALA & -79.94 (0.1)& -115.84 (0.1) & -116.93 (0.3)& 7.209 (0.005)\\
VAE-dsMALA    & -80.36 (0.1)& -116.32 (0.2)      & -117.48 (0.2)& 7.203 (0.001)   \\
VAE-gradHMC    & \textbf{-79.52} (0.2)& \textbf{-115.77 }(0.1) & \textbf{-116.69 }(0.3)& \textbf{7.179} (0.001) \\
VAE-dsHMC     & -79.89 (0.1)&  -116.02 (0.1) & -116.88 (0.1) &  7.187 (0.003) \\
\hline
VAE-MoG      & -80.52 (0.1)& -116.40 (0.3)  & -119.14 (0.1)&  7.205 (0.001) \\
VAE-VAMP      & \textbf{-78.48} (0.1)& \textbf{-114.30} (0.1) & \textbf{-117.23 }(0.1)&\textbf{7.197} (0.001)        \\
\hline
\end{tabular}
\end{sc}
\end{small}
\end{center}
\vskip -0.1in
\end{table*}

\paragraph{Marginal log-likelihood estimation.} We start to evaluate the performance of different variations of VAEs using the marginal log-likelihood of the model on a held-out test set for a variety of benchmark datasets. In doing so, we resort to importance sampling to estimate the marginal log-likelihood using $S$ importance samples via
$$\log \hat{p}_{\text{IS}} (x) = \log \frac{1}{S}\sum_{s=1}^{S}\frac{p_\theta(x|z_{s})p_\theta(z_{s})}{r(z_{s}|x)} ~, z_s \sim r(\cdot|x),$$
where $r$ is an importance sampling density. Following \citet{ruiz2019contrastive}, in the case of a standard VAE, we choose $r(z|x)=\mcn(\mu_{\phi_0}^z(x), \tau \Sigma_{\phi_0}^z(x))$ for some scaling constant $\tau\geq 1$, assuming that $q^0_{\phi_0}(z_0|x)=\mcn(\mu_{\phi_0}^z(x), \tau \Sigma_{\phi_0}^z(x))$ with diagonal $\Sigma_{\phi_0}^z(x)$. For the case with MCMC sampling using $K$ steps, we choose $r(z_s|x)=  \mcn(z_K(x), \tau\Sigma_{\phi_0}^z(x))$, where $z_K(x)$ is an estimate of the posterior mean from the MCMC chain.




\paragraph{VAE models.} Using the metric described above, we evaluate our model and compare it against other popular adjustments of VAEs for various data sets. In terms of comparing models, we focused on comparing our model against a Vanilla VAE, MCMC coupled VAE using a dual-averaging adaptation scheme \citep{hoffman2014no, nesterov2009primal}, and VAEs using more expressive priors such as a Mixture of Gaussians (MoG), cf. \citet{jiang2017variational, dilokthanakul2016deep}, or a Variational Mixture of Posteriors Prior (VAMP), see \citet{tomczak2017vae}. 
For the MNIST example, we consider a Bernoulli likelihood with a latent dimension of size 10. We pretrain the model for 90 epochs with a stanadard VAE, and subsequently trained the mode for 10 epochs with MCMC. We used a learning rate of 0.001 for both algorithms. For the remaining datasets, we pre-trained for 290 epochs with a standard VAE, followed by training for 10 epochs with MCMC. We used a learning rate of 0.005, while the latent dimension size is 10, 20, and 64 for Fashion-MNIST, Omniglot and SVHN, respectively. For the the SVHN dataset, we considered a 256-logistic likelihood with a variance fixed at $\sigma^2 = 0.1$. In terms of the neural network architecture used for the encoder and the decoder, more information can be found in the code-base, 


\paragraph{Experimental results.} Table \ref{log-likelihood-vae} summarizes the estimated log-likelihoods for the different data sets. The results therein show the means of three independent runs, with their standard deviation in brackets. For the case of SVHN, the estimate is transformed to be represented in bits per dimension. We observe that among the considered methods utilizing MCMC for VAEs, our approach performs better across the datasets we explored. We note that for the decoder and encoder models considered here, the use of different generative models with more flexible priors such as the VAMP prior can yield higher log-likelihoods. However, the choice of more flexible priors is completely complementary to the inference approach suggested in this work. Indeed, we illustrate in Sections \ref{sec:linear_hvae_experiments} and \ref{sec:linear_hvae_experiments} that our MCMC adaptation strategy performs well for more flexible hierarchical priors.

\subsection{Evaluating Generative Performance with Kernel Inception Distance (KID)}

\paragraph{Generative metrics.} The generative performance of our proposed model is additionally quantitatively assessed by computing the Kernel Inception Distance (KID) relative to a subset of the ground truth data.We chose the KID score instead of the more traditional Fréchet inception distance (FID), due to the inherent bias of the FID estimator \citep{binkowski2018demystifying}. To compute the KID score, for each image from a held-out test set given a particular dataset, we sample a latent variable from the prior density and then pass it through the trained decoder of the corresponding model to generate a synthetic image. Images are resized to (150,150,3) using the bi-cubic method, followed by a forward pass through an inceptionV3 model using the Imagenet weights. This yields a set of Inception features for the synthetic and held-out test set. The computation of the KID score for these features is based on a polynomial kernel, similarly to \citet{binkowski2018demystifying}. For all datasets, we utilized a learning rate of 0.001 for both the VAE and MCMC algorithms. We trained the VAE for 100 epochs and performed sampling with the MCMC algorithms for 50 epochs if applicable, yielding a total training of 150 epochs across all cases. The likelihood functions used was Bernoulli for the MNIST and Fashion-MNIST datasets, while the logistic-256 was used for the SVHN and Cifar-10 datasets, with a fixed variance of $\sigma^2 = 0.1$ and $\sigma^2 = 0.05$, respectively. The dimension of the latent variable was fixed to 10 for the MNIST datasets, while it was set to 64 and 256 for the SVHN and CIFAR-10 datasets. More details regarding the neural network architecture used for training the VAE can be found in the codebase. 

\begin{table*}[t]
\caption{Estimates of KID for each model considered across different datasets.The values denote the mean of three seeds, while the standard deviation is given within brackets. The MoG and VAMP VAEs use different priors.}
\label{kid-score-vae}
\vskip 0.15in
\begin{center}
\begin{small}
\begin{sc}
\begin{tabular}{lcccr}
\hline
Model & MNIST  & Fashion-MNIST & SVHN & CIFAR-10\\
\hline
VAE    & 1.084 (0.05) & 0.925 (0.03) & 0.197 (0.01) & 1.348 (0.01) \\
VAE-gradHMC    & \textbf{0.431} (0.02) & \textbf{0.852 }(0.03) & \textbf{0.126} (0.01) & \textbf{1.153} (0.01) \\
VAE-dsHMC     & 0.653 (0.01)&  0.908 (0.06) & 0.183 (0.02)&  1.587 (0.10) \\
\hline
VAE-MoG      & 0.542 (0.01)& 0.990 (0.01) & \textbf{0.190 }(0.01) &  \textbf{1.444 }(0.01)   \\
VAE-VAMP      & \textbf{0.434 }(0.05)& \textbf{0.610 }(0.03) & 0.210 (0.01) &1.657 (0.03)         \\
\hline
\end{tabular}
\end{sc}
\end{small}
\end{center}
\vskip -0.1in
\end{table*}

\paragraph{VAE models and quantitative evaluation.} Similarly to Section 5.1, we perform a series of experiments comparing our model to other popular VAE adaptations across different data sets. 
In Table \ref{kid-score-vae}, we summarize the results of our experiments reporting mean KID scores from three different seeds with the standard deviation in brackets. We notice a similar pattern to that in Section 5.1 where our proposed method outperforms other MCMC-related methods. At the same time, we observe that models with more expressive prior such as the VAMP prior can perform equally or slightly better in the case where the latent dimension is small, such as for MNIST and Fashion-MNIST. However, in the case of a higher dimensional latent space such as CIFAR-10 with $d_z=256$, we observe that our method shows considerable improvement compared to the other methods being tested.

\begin{figure}[ht]
\vskip 0.2in
\begin{center}
\centerline{\includegraphics[width=\columnwidth]{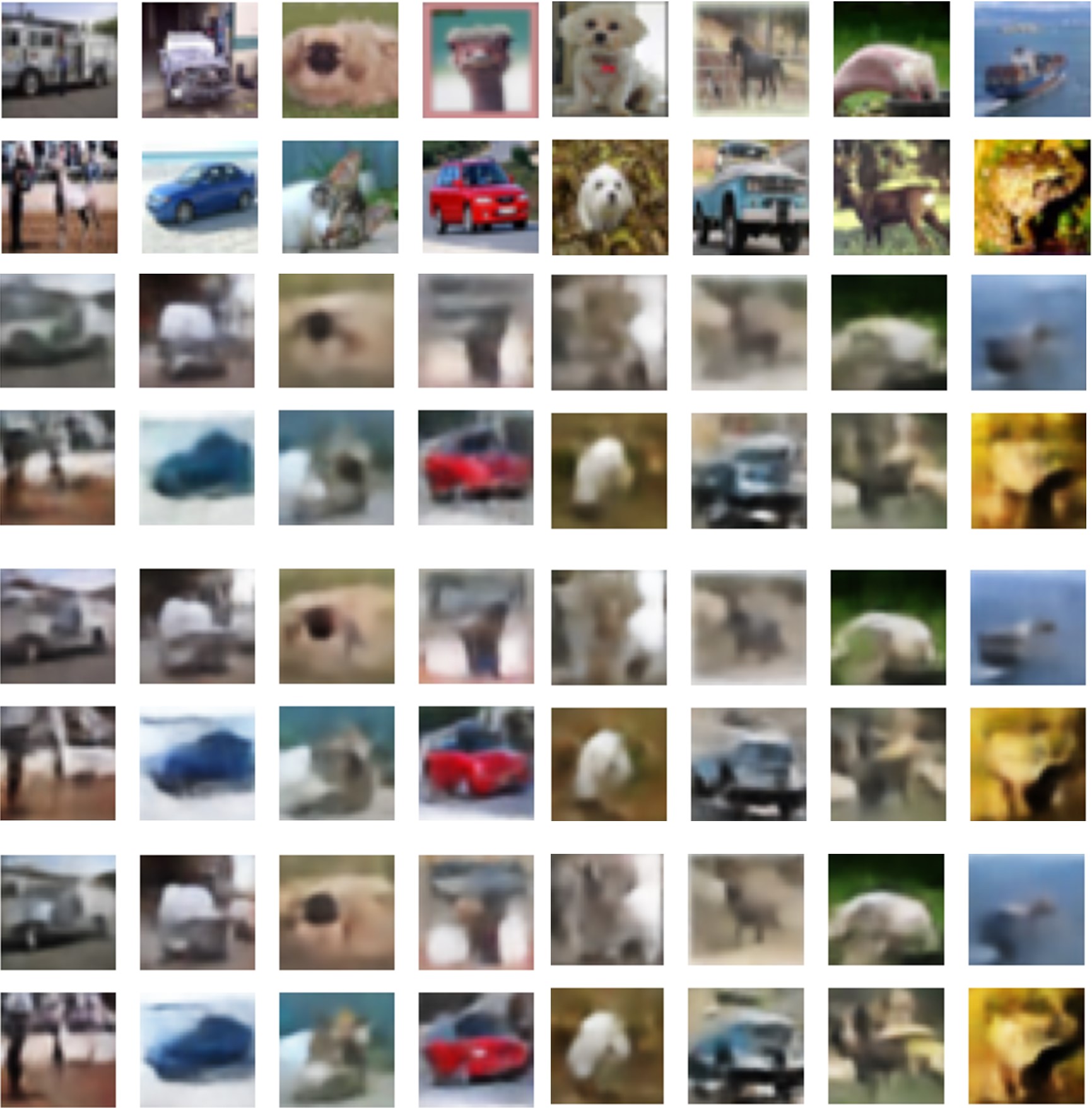}}
\caption{Model reconstruction images for the top three performing models tested on CIFAR-10 in terms of the KID-score evaluated on model samples. The first two rows illustrate the ground truth followed by the next two rows showing reconstructions from the Vanilla VAE model followed by the next two rows illustrating reconstructions from the dsHMC model, and finally, the last two rows illustrating reconstructions from the gradHMC coupled VAE.}
\label{model-reconstructions}
\end{center}
\vskip -0.2in

\end{figure}

\begin{figure}[ht]
\vskip 0.2in
\begin{center}
\centerline{\includegraphics[width=\columnwidth]{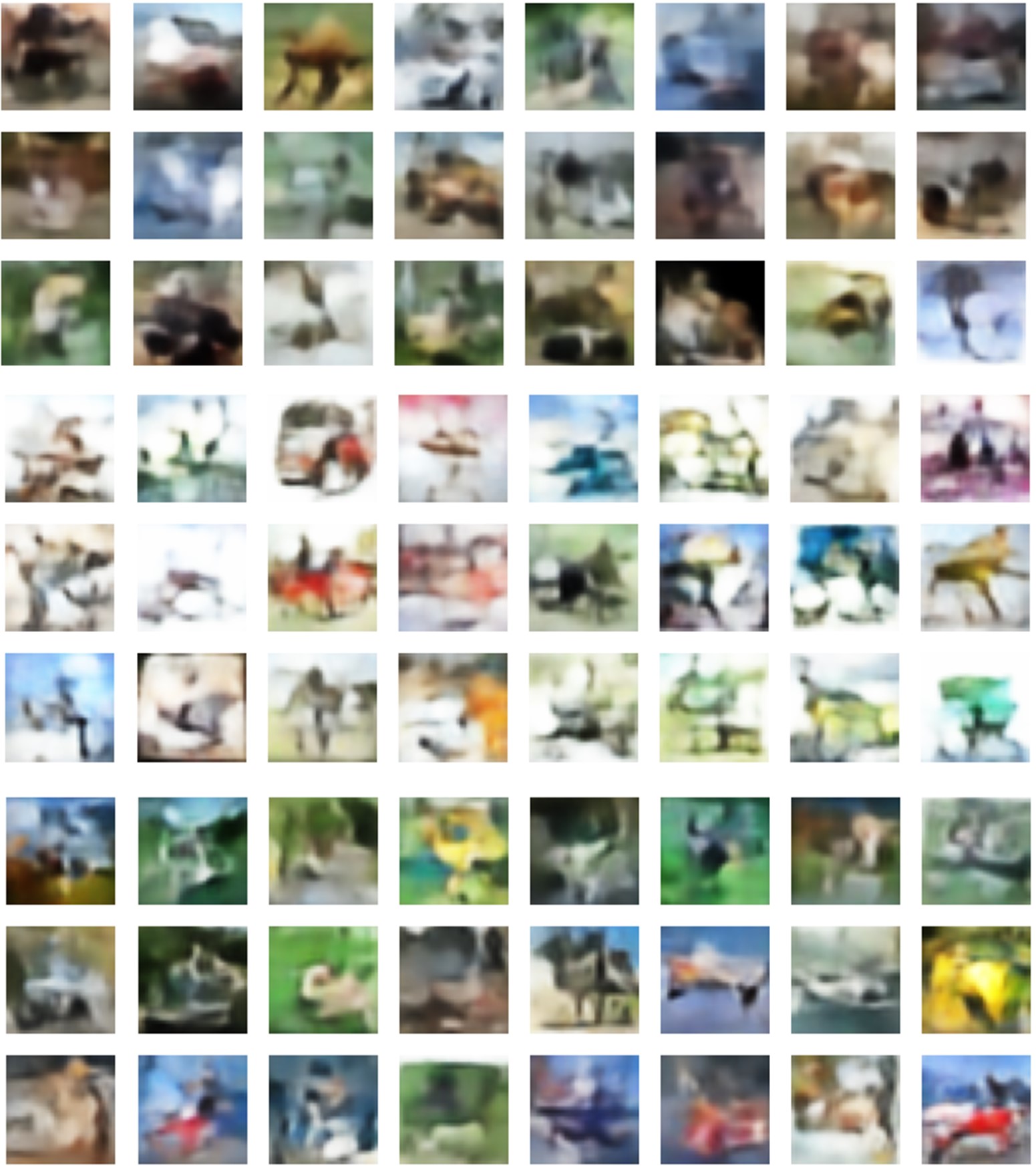}}
\caption{Model sampled images for the top three performing models tested on CIFAR-10 in terms of the KID-score evaluated on model samples. The first three rows illustrate samples from the Vanilla VAE model followed by the next three rows illustrating samples from the dsHMC model, and finally the last three rows illustrating samples from the gradHMC coupled VAE.}
\label{model-samples}
\end{center}
\vskip -0.2in
\end{figure}

\paragraph{Qualitative results.} In addition to computing the KID score, we qualitatively inspect the reconstructed images and the images sampled from the model, as used for the computation of the KID score in the section above. In Figure \ref{model-reconstructions}, we can see reconstruction for the best three performing models, while in Figure \ref{model-samples}, we can see unconditionally generated samples for the same models. We observe that, indeed, KID score qualitatively correlates with more expressive generations and reconstructions. In particular, we observe a slight decrease in blurriness and an increase in the resolution of smaller details such as the car-light of the red car in Figure \ref{model-reconstructions}. Moreover, the unconditionally generated images in Figure \ref{model-samples} exhibit more expressive color patterns.

\subsection{Evaluating Model Performance in Small Sample Size Data}

\paragraph{Data augmentation task.} In addition to testing our proposed model against the benchmarks in the previous section, we also test our model in a real-world dataset comprised of complex images that, however, has a relatively small sample size. We chose the Alzheimer’s Disease Neuroimaging Initiative (ADNI) 
\footnote{https://www.kaggle.com/datasets/sachinkumar413/alzheimer-mri-dataset} brain MRI dataset, which is comprised of 4000 Brain MRI Scans of individuals suffering from Dementia, and individuals from a healthy control group in a ratio of 1:3, respectively. The small sample size as well as the imbalance in the dataset pose a problem for classification tasks that are often addressed by different data augmentations. We illustrate here that the proposed generative model can be used to generate additional samples that are representative of each population in the dataset, namely healthy controls and diseased individuals. We first trained VAEs with each separate class on the dataset, which we modified by fusing the dementia classes into a single class compared two using three separate dementia classes using a VAE learning rate of 0.001 and a MCMC learning rate of 0.01, where it was applicable.

\begin{figure}[ht]
\vskip 0.2in
\begin{center}
\centerline{\includegraphics[width=\columnwidth,scale=0.25]{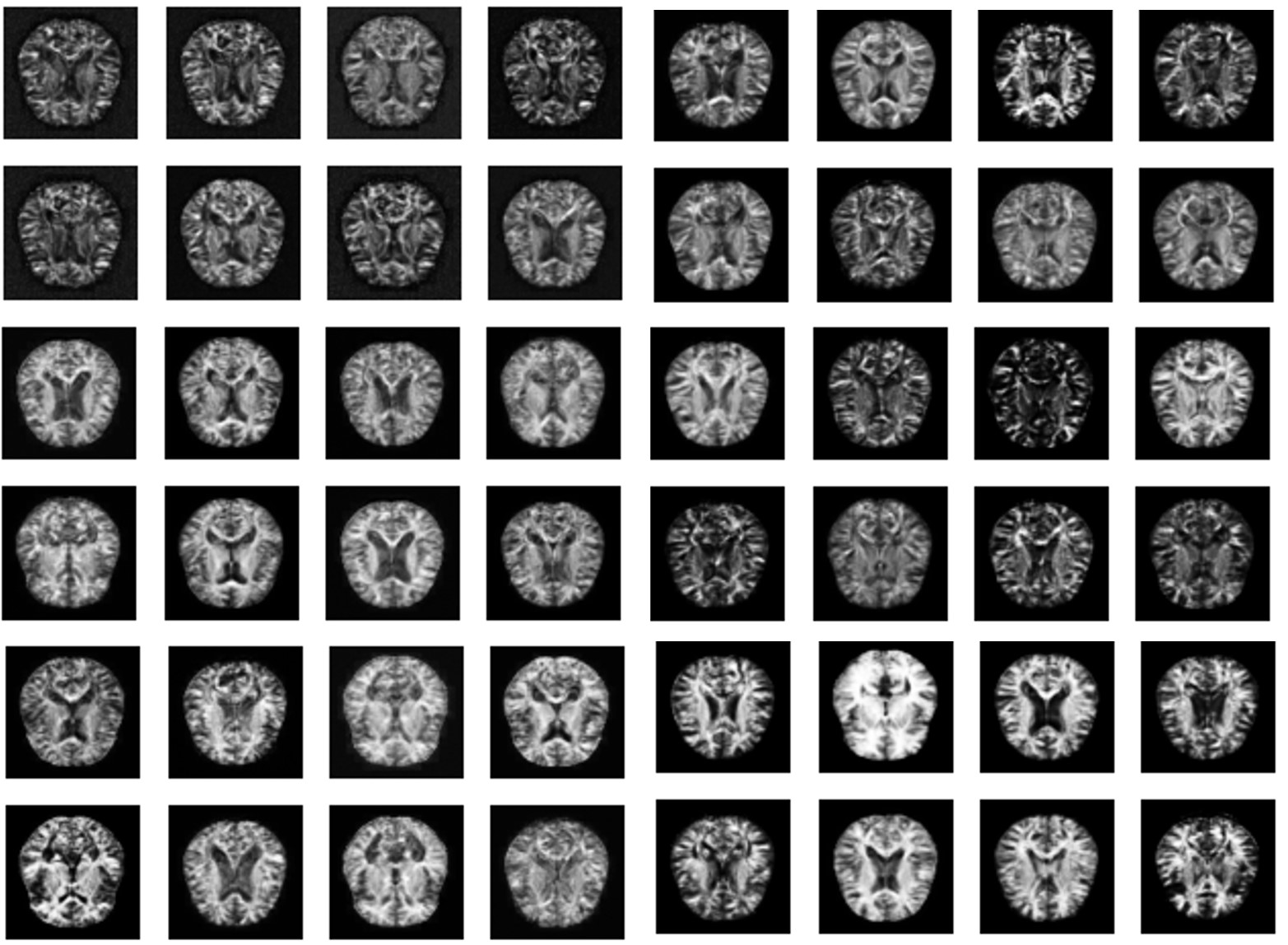}}
\caption{Model samples from VAE variations trained on either demented (first four columns) or normal patients (last four columns). The first two rows are samples from the Vanilla VAE model, the next two rows from the VAE using dual-average adaptation and the last two rows from our proposed method using a VAE with entropy based adaptation}
\label{brain-model-samples}
\end{center}
\vskip -0.2in
\end{figure}

\paragraph{Generative performance.}
The VAEs were trained for 2000 epochs with 100 epochs of MCMC coupled training, where it was applicable. The KID score presented was formed by comparing the whole dataset (that is, including both training and test sets) to the same dimension of models samples, due to the KID score underperforming for the small size of the test set for the minority class. The neural networks utilized in the encoder and the decoder were similar to those of Section 7.1, consisting of two dense layers of 200 units each for the decoder and the encoder. Moreover, the latent dimension for all experiments was fixed at 20, while the likelihood utilized was a logistic-256 with fixed variance of $\sigma^2 = 0.05$. After training, a series of 200, 500, 1000, and 2000 images were generated for the minority class, which were then augmented with the generated images. Classification performance for classifier models trained on this augmented dataset was then compared against classifier models trained on the non augmented dataset. More details regarding the architectures used for the VAE and classifier models can be found in the codebase. We observed that one obtains the best performance in terms of the classification metrics for the dataset augmented with 200 images and thus we report these values in \ref{kid-score-vae-adni}. We find that a VAE with a gradient-based HMC sampler has better generative performance, particularly for the dementia group. The minority class, \ie~ the dementia group, was augmented by the addition of synthetic data from the generative models. Qualitiative results showing the generated samples are given in Figure \ref{brain-model-samples} for the standard VAE model, and those combined with MCMC, either adapted with a dual average adaptation or an entropy based adaptation. We notice that our proposed method captures more brain characteristics for both the demented and normal patients, due to presence of various brain structures throughout the generated samples, while also capturing class specific characteristics, such as a greater degree of brain matter loss in the dementia class. 

\begin{table*}[htb]
\caption{Estimates of the KID score for each respective class in the ADNI brain MRI dataset and classification metrics from the data augmentation task across different models. Standard deviations in brackets.}
\label{kid-score-vae-adni}
\vskip 0.15in
\begin{center}
\begin{small}
\begin{sc}
\begin{tabular}{lccccc}
\hline
Model & KID/Dementia & KID/Controls & Bacc & TPR & TNR\\
\hline
VAE    & 12.44 (0.8) & 12.64 (2.35) & 0.968 (0.01) & 0.986 (0.003) & 0.950 (0.002) \\
VAE-gradHMC & \textbf{10.25} (0.85)  & \textbf{9.76} (1.11)& \textbf{0.971} (0.01)  & \textbf{0.989} (0.005) & \textbf{0.954} (0.002) \\
VAE-dsHMC   &  12.02 (1.67) & 10.81 (1.28)   & 0.964 (0.01) & \textbf{0.989} (0.005) & 0.940 (0.001) \\
No-Augmentation & - & - & 0.878 (0.02) & 0.824 (0.024) & 0.932 (0.025)  \\
\hline
\end{tabular}
\end{sc}
\end{small}
\end{center}
\vskip -0.1in
\end{table*}

\paragraph{Classification results.}

We performed a classification between the two groups with results summarised in Table \ref{kid-score-vae-adni}, showing first, that augmenting data with a trained VAE improves the classification in general, and second, that augmentations with our proposed method lead to a small, yet significant increase in Balanced Accuracy, True Positive Rate (TPR) and True Negative Rate (TNR). These results are consistent with the improved quality of the generated samples using our approach and we thus believe that our method can be leveraged for effective data augmentations.

\subsection{Linear hierarchical VAEs}
\label{sec:linear_hvae_experiments}
We consider linear Gaussian models with a Gaussian prior $p_{\theta}(z)=\mcn(\mu_z,\Sigma_z)$ and a linear decoder mapping so that $p_{\theta}(x|z)=\mcn(Wz+b, \Sigma_{x|z})$ for $\mu_z, b \in \rset^{d_z}$, $W\in \rset^{d_x \times d_z}$ and covariance matrices $\Sigma_z$ and $\Sigma_{x|z}$ of appropriate dimension. The resulting generative model corresponds to a probabilistic PCA model \citep{tipping1999probabilistic}, see also \citet{dai2018connections,lucas2019don} for further connections to VAEs. The aim of this section is to illustrate that adaptation with a non-diagonal pre-conditioning matrix becomes beneficial to account for the dependence structure of the latent variables prevalent in such hierarchical models.


\paragraph{Hierarchical generative model.} We can sample from the Gaussian prior $z\sim \mcn(\mu_z, \Sigma_z)$ in a hierarchical representation using two layers: 
\begin{equation}
    z_1 \sim \mcn(0, \I), \quad z_2|z_1 \sim \mcn(A_2z_1+c_2^{\mu}, \Lambda_{z_2|z_1})) \label{eq:two_layer_gaussian_prior},
\end{equation} 
where $z=(z_1,z_2)$ and $\Lambda_{z_2|z_1}=\text{diag}(\sigma_{z_2|z_1}^2)$. To recover \eqref{eq:two_layer_gaussian_prior} from the general auto-regressive prior factorisation \eqref{eq:prior_hvae}, assume that $d_1=0 \in \rset^{n_1'}$, $n_1'=n_1'$. For $d=(d^{\mu},d^{\sigma})$, suppose that $\mu_{1,\theta}(d)=d^{\mu}$ is the projection on the first $n_1$ components while $\sigma_{1,\theta}(d)=\exp(0.5 d^{\sigma})$ describes the standard deviation based on the last $n_1$ components. Further, consider the linear top-down mapping $$h_{2, \theta}\colon (z_1, d_1) \mapsto d_2=\begin{bmatrix} A_2 & B_2 \\ 0 & 0 \end{bmatrix} \begin{bmatrix} z_1 \\ d_1 
\end{bmatrix} + \begin{bmatrix} c_2^\mu \\ c_2^\sigma \end{bmatrix}, $$ 
for the deterministic variables, where
$c_2^\sigma =2 \log \sigma_{z_2|z_1}$. We assume the same parameterisation for the prior densities of $z_2$ given  $d_2$ as in the first layer:  $\mu_{2,\theta}(d)=\mu_{1,\theta}(d)=d^\mu$, and $\sigma_{2,\theta}(d)=\sigma_{1, \theta}(d)=\exp(0.5 d^{\sigma})$.
We assume further that the decoder function depends explicitly only on the latent variables $z_2$ and $d_2$ at the bottom in the form of 
\begin{align*}
p_{\theta}(x|z)=&\mcn(W_2^z z_2 +W_2^d d_2 +b, \Sigma_{x|z})\\=& \mcn(W z+b+W_2^d c_2^{\mu}, \Sigma_{x|z}),
\end{align*}
for $W=\begin{bmatrix} W_2^d A_2 & W^z_2 \end{bmatrix}$.
Observe that the covariance matrix of the prior density is
$$ \Sigma_z = \begin{bmatrix} \I & (A_2)^\top \\ A_2  &  A_2  A_2 ^\top + \I \end{bmatrix}.$$
The marginal distribution of the data is
$x \sim \mcn(\mu_x, \Sigma_x)$ where
$\mu_x=W_2^z c_2^{\mu} + b$ and $$\Sigma_x=W \Sigma_z W^\top + \Sigma_{x|z}.$$
The covariance matrix of the posterior density becomes
\begin{equation}
    \Sigma_{z|x} = \Sigma_z - (W\Sigma_z)^\top \Sigma_x^{-1} W\Sigma_z, \label{eq:posterior_gaussian}
\end{equation}
which can be badly conditioned so that MCMC methods without pre-conditioning can perform poorly. However, if we can learn a pre-conditioning matrix $C$ such that $C^\top \Sigma_{z|x}^{-1} C$ becomes well conditioned, then MALA or HMC can be potentially more effective.

\paragraph{Encoding model.}
Assume a linear encoder model based on a linear bottom-up model so that $d_3'=x$ and for $1 \leq \ell \leq 2$, suppose that $d_{\ell}'=W_{\ell}'d'_{\ell+1}+b_{\ell}'$ are bottom-up deterministic variables. 
We construct an encoding distribution by setting 
$$ \mu'_{\ell,\theta} \colon (d_{\ell},d_{\ell}') \mapsto B'_{\ell} \begin{bmatrix} d_{\ell} \\d_{\ell}' \end{bmatrix} + c_{\ell}' $$
and $ \sigma'_{\ell,\theta}\colon (d_{\ell},d_{\ell}') \mapsto \exp(b_{\ell}')$ in the residual parameterization \eqref{eq:residual_param}.

\begin{table*}[t]
\caption{Computation of the Condition Number of the posterior for a linear Hierarchical VAE with two layers of dimensionality (10,20) and (50,100), along with the condition number of the transformed posterior.}
\label{chvae-condition-number}
\vskip 0.15in
\begin{center}
\begin{small}
\begin{sc}
\begin{tabular}{lcccr}
\hline
Model & $\kappa(\Sigma_{z|x}^{-1})_{(10,20)}$  & $\kappa(C^\top \Sigma_{z|x}^{-1} C)_{(10,20)}$& $\kappa(\Sigma_{z|x}^{-1})_{(50,100)}$  & $\kappa(C^\top \Sigma_{z|x}^{-1} C)_{(50,100)}$ \\
\hline
hVAE    &  18.07 (0.45) & - &  508.55 (12.1) & - \\
gradMALA-D & 20.43 (0.99)& 19.02 (1.24) & 578.05 (84.5)& 434.24 (13.0)\\
dsMALA-D    & 18.62 (1.34)& 18.62 (1.34) & 617.36 (48.4)& 617.36 (48.4)     \\
gradHMC-D    & 21.57 (1.48)& 22.67 (1.23) & 502.41 (36.0)& 431.14 (23.4) \\
dsHMC-D    & 18.38 (1.27)&  18.38 (1.27)& 621.1 (54.9)&  621.1 (54.9)\\
gradMALA-LT     & 23.55 (5.31)& \textbf{1.67 }(0.06)& 475.93 (23.7)& \textbf{2.0 }(0.02) \\
gradHMC-LT    & 25.57 (2.63)& 1.68 (0.15)& 483.05 (38.0)& 2.24 (0.04)         \\
\hline
\end{tabular}
\end{sc}
\end{small}
\end{center}
\vskip -0.1in
\end{table*}

\paragraph{Experimental results.}

We first test if the adaptation scheme can adapt to the  posterior covariance $\Sigma_{x|z}$ given in \eqref{eq:posterior_gaussian} of a linear hVAE model, \ie~ if the condition number of $C\Sigma_{x|z}C^\top$ becomes small. As choices of $C$, we consider (i) a diagonal preconditioning matrix (denoted D) and (ii) a lower-triangular preconditioning matrix (denoted LT). Note that the dual-averaging adaptation scheme used here and in \cite{hoffman2014no} adapts only a single step-size parameter, thereby leaving the condition number unchanged. We tested two simulated data sets with corresponding latent dimensions $(n_1,n_2)$ of (10,20) and (50,100). More specifically, we simulated datasets with 1000 samples for each condiguration, using the linear observation model with a standard deviation of 0.5. We used a hierarchical VAE with two layers and a learning rate of 0.001. 
For the dataset from the model with a latent dimension of (10,20), we pre-trained the vVAE for 1000 epochs without MCMC, followed by training for 1000 epochs with MCMC. The
the number of MCMC steps was fixed at $K=2$. 
For the dataset from the model generated from a higher dimensional latent space of dimension (50,100), we increased the number of training epochs from 1000 to 5000, while also increasing the number of MCMC steps from  $K=2$ to $K=10$. For different choices for the size of the latent variables, Table \ref{chvae-condition-number} shows that both gradient-based adaptation schemes lead to a very small transformed condition number $\kappa(C^\top \Sigma_{z|x}^{-1} C)$ when a full preconditioning matrix is learnt. Notice also that for all models, the posterior becomes increasingly ill-conditioned for higher dimensional latent variables, as confirmed by the large values of $\kappa(\Sigma_{z|x}^{-1})$ in Table \ref{chvae-condition-number}.

\begin{table}[htb]
\caption{Difference between true and estimated data log-likelihood $\log p_{\theta}(x)$ for hierarchical VAEs with two layers and (10,20), (50,100) latent dimensionality respectively.}
\label{chvae-log-likelihood}
\vskip 0.15in
\begin{center}
\begin{small}
\begin{sc}
\begin{tabular}{lcccr}
\hline
Model & $\Delta \log p(x)_{(10,20)}$   & $\Delta \log p(x)_{(50,100)}$\\
\hline
hVAE    &  24.91 (7.67) & 13.08 (0.26) \\
gradMALA-D & 1.54 (1.49)& 7.14 (0.11)\\
dsMALA-D    & 2.68 (2.04)& 8.77 (0.11)         \\
gradHMC-D    & 1.16 (1.72)& 2.19 (0.05) \\
dsHMC-D    & 2.59 (1.78)&  8.06 (0.48)\\
gradMALA-LT     & 1.56 (1.61)& 1.96 (0.12) \\
gradHMC-LT    & \textbf{1.14} (1.53)& \textbf{1.66} (0.15) \\
\hline
\end{tabular}
\end{sc}
\end{small}
\end{center}
\vskip -0.1in
\end{table}

In addition to the condition number, we also investigate how the adaptation scheme affects the learned model in terms of the marginal log-likelihood, which is analytically tractable. The results summarized in Table \ref{chvae-log-likelihood} show that the gradient-based adaptation schemes indeed achieve a higher log-likelihood.


\subsection{Non-linear hierarchical VAEs}
\label{sec:non_linear_hvae_experiments}

Finally, we investigate the effect of combining MCMC with hVAE in the general non-linear case for hierarchical models. We consider a hVAE with two layers of size $5$ and $10$. The learning rate of the hVAE and MCMC algorithms was set to 0.001. We use 200 epochs for training overall. For models that included MCMC sampling, we used the first 190 epochs for pre-training without MCMC. Additionally, the prior of the model was trained only during the hVAE portion of the algorithm. The resulting KID scores for MNIST and Fashion-MNIST can be found in Table \ref{non-linear-chvae-kid-score}. In this scenario, our proposed method outperforms other sampling schemes when combined with a hVAE model. 

\begin{table}[htb]
\caption{Estimates of KID for each model considered across different datasets.The values denote the mean of three seeds, while the standard deviation is given within brackets.}
\label{non-linear-chvae-kid-score}
\vskip 0.15in
\begin{center}
\begin{small}
\begin{sc}
\begin{tabular}{lcccr}
\hline
Model & MNIST  & Fashion-MNIST \\
\hline
hVAE    &  0.496 (0.062) & 1.269 (0.037) \\
gradMALA-D & 0.432 (0.057)& 1.038 (0.068)\\
dsMALA-D    & 0.600 (0.040)& 1.312 (0.034)         \\
gradHMC-D    & 0.447 (0.073  )& 1.079 (0.174) \\
dsHMC-D    & 0.490 (0.098)&  1.151 (0.223)\\
gradMALA-LT     & 0.475 (0.101)& 0.939 (0.143) \\
gradHMC-LT    & \textbf{0.407} (0.030)& \textbf{0.916} (0.047)         \\
\hline
\end{tabular}
\end{sc}
\end{small}
\end{center}
\vskip -0.1in
\end{table}


\section{Conclusion}

We have investigated the performance effect of training VAEs and hierarchical VAEs with MCMC speed measures and subsequently compared our proposed method with other widely used adaptive MCMC adaptations and VAE model variations. 
Adopting recent advances in adaptive MCMC literature that are based on the notion of a generalised speed measure 
seems to provide, in the problems and datasets we tested, a more efficient learning algorithm for VAEs.
Future research directions may focus on using our proposed method in models with deeper architectures in the encoder and the decoder, using our method in challenging inpainting problems and exploring its power at alleviating adversarial attacks as seen in \citet{kuzinaalleviating}.

\newpage
\bibliography{bibliography,bib2}

\end{document}